\newcommand\DSTwoTrainCountAll{4167}
\newcommand\DSTwoTrainCountDirected{3325}
\newcommand\DSTwoTrainCountRound{2055}
\newcommand\DSTwoTrainCountInstancesAll{5380}

\PassOptionsToPackage{table,xcdraw}{xcolor}
\documentclass{article}
\usepackage{colortbl}
\usepackage{algorithmic}
\usepackage{graphicx}
\usepackage{url}
\graphicspath{{C:/Article_figs/},{./fig/}}
\usepackage{textcomp}
\usepackage{tabularx,booktabs}
\usepackage{multirow}
\usepackage{array}
\usepackage{flafter}
\usepackage{subcaption}
\usepackage{spconf}

\usepackage{compactbib}

\usepackage{enumitem}

\usepackage{hyperref}

\usepackage{comment}
\usepackage{balance}

\newcommand{\specialcell}[2][c]{%
	\begin{tabular}[#1]{@{}c@{}}#2\end{tabular}}

\usepackage{color}
\usepackage{framed}

\newcounter{t0d0_counter}
\newcommand{\nofixme}[1]{
}

\usepackage{csvsimple}

\def\BibTeX{{\rm B\kern-.05em{\sc i\kern-.025em b}\kern-.08em
    T\kern-.1667em\lower.7ex\hbox{E}\kern-.125emX}}

\title{\texttt{BRIMA}: low-overhead \textit{Browser-only Image Annotation} tool (Preprint)}

\name{Tuomo Lahtinen$^{\dagger}$, Hannu Turtiainen$^{\ddagger}$, Andrei Costin$^{\ddagger}$\sthanks{To appear in \emph{Proceedings of the 2021 IEEE International Conference on Image Processing (ICIP).}}}
\address{University of Jyv\"askyl\"a \\ 
Jyv\"askyl\"a, Finland \\ $^{\dagger}$ tuomo.t.lahtinen@student.jyu.fi , $^{\ddagger}$ \{ancostin,turthzu\}@jyu.fi }

\begin{document}

\maketitle



\begin{abstract}

Image annotation and large annotated datasets are crucial parts within 
the Computer Vision and Artificial Intelligence fields.
At the same time, it is well-known and acknowledged by the research community 
that the image annotation process is challenging, time-consuming and hard to scale. 
Therefore, the researchers and practitioners are always seeking ways to 
perform the annotations easier, faster, and at higher quality. 
Even though several widely used tools exist and the tools' landscape evolved 
considerably, most of the tools still require intricate technical setups and 
high levels of technical savviness from its operators and crowdsource contributors. 

In order to address such challenges, we develop and present 
\texttt{BRIMA} -- a flexible and open-source browser extension that 
allows \textbf{BR}owser-only \textbf{IM}age \textbf{A}nnotation at considerably lower overheads. 
Once added to the browser, it instantly allows the user to annotate images easily 
and efficiently directly from the browser without any installation or setup 
on the client-side. It also features cross-browser and cross-platform 
functionality thus presenting itself as a neat tool for researchers 
within the Computer Vision, Artificial Intelligence, and privacy-related fields. 

\end{abstract}

\section{Introduction}
\label{sec:intro}

Research efforts in the fields of Computer Vision, image processing, 
image annotation, and object detection/recognition often require large datasets 
of annotated images. 
In some cases, existing datasets are not 
suitable or not available, therefore the data must be created fast and 
from the scratch as part of the research. 
%
For image dataset creation purpose, multiple annotation tools have been 
developed and published during the last 
decade~\cite{ciocca2015iat,halaschek2005photostuff,bernal2019gtcreator,russell_labelme,wada_labelme,labelimg,korc2007annotation}. 
In order to create such an object detection model, it requires 
a large and high-quality dataset of annotated images. Therefore, there is 
stringent need for fast, easy and scalable image annotation tools with 
lowest overheads in terms of bootstrapping and usage. 

This work presents the first browser-only image annotation tool which 
aids researchers to fast and easy create high-quality image-based 
datasets for their custom needs. 
Our browser extension offers a new way to annotate images, where annotation 
is done directly in the browser (within browser-viewport), does not require additional software 
setups, and the annotator is in full control which image to annotate 
(as it comes directly from the browser's viewport), allowing what could be called 
\emph{annotate-while-freely-browsing} experience. 
The support for standard/compatible 
data formats (e.g., MS COCO~\cite{lin2014microsoft}) makes our extension 
compatible with other annotation tools that are used in the computer vision 
modeling pipelines.

For the browser extension, we set forward a few requirements to ease our collaborative 
annotation team, in making image annotation with good quality and coverage. 
Firstly, the extension should be easy and straightforward to use.
For example, the user does not need to learn how to use extension more 
than a few minutes and annotation should be a forward 
process where the UI of the extension makes user comfortable to perform 
the annotations at high pace and with high accuracy while offering a 
good User Experience (UX). 
%
Secondly, the extension should require minimal setup. 
For example, the most recent browsers are known for the ease of 
installing add-on software, i.e., browser extensions. It is as simple as 
clicking ``Install Add-on'' and not even requiring the browser's restart, 
all that thanks to robust browser architectures and the great flexibility 
of high-level programming languages such as JavaScript. 
%

Our main contributions with this work are as follow:
\begin{itemize}[leftmargin=*]
\itemsep0em

\item To the best of our knowledge, we develop the first browser-only browser-viewport image annotation toolset 

\item We validate our tool's effectiveness in a real-world crowdsourcing experiment

\item We release the code and artifacts under open-source license 


\vspace{-4mm}
\end{itemize}


\section{Related work}
\label{sec:related}

%
PhotoStuff is an \emph{``annotation tool for digital images on the Semantic Web''} 
allowing \emph{``to manually annotate images using Web ontologies''} proposed by Halaschek et al.~\cite{halaschek2005photostuff}.
Presumably, it could be run both as a stand-alone as well as a web-based application. 
Wilhelm et al.~\cite{wilhelm2004photo} presented photo annotation with 
the mobile phone. In their work, some annotation was created automatically 
and sent to a server for annotation enhancements.
%
LabelMe by Russell et al.~\cite{russell_labelme} is an earlier web-based 
image annotation tool, that allowed pretty flexible annotation and functionality. 
%
Korc and Schneider~\cite{korc2007annotation} proposed MATLAB-based Annotation 
Tool -- a stand-alone software employing parts of the LabelMe by Russell et al.~\cite{russell_labelme}. 

LabelImg~\cite{labelimg} is a tool that presents a fast way to perform rectangle 
annotation to images in PascalVOC~\cite{eve_voc} and YOLO~\cite{redmon_yolo3} formats, 
but it does not support MS COCO~\cite{lin2014microsoft} JSON format. 
%
Labelme by Wada~\cite{wada_labelme} can annotate images with polygon segments 
that can be subsequently used in object segmentation architectures. 
It also supports MS COCO~\cite{lin2014microsoft} JSON format, and includes 
Python scripts to embed the data as an MS COCO single annotation file format. 
Labelme by Wada~\cite{wada_labelme} has drawn inspiration from LabelMe 
by Russell~\cite{russell_labelme}. 
%
Ilastik is a stand-alone \emph{``easy-to-use interactive tool that 
brings machine-learning-based (bio) image analysis to end users without 
substantial computational expertise''} developed and proposed by 
Sommer et al.~\cite{sommer2011ilastik} and Berg et al.~\cite{berg2019ilastik}. 
%
Ciocca et al.~\cite{ciocca2015iat} developed IAT as a stand-alone annotation tool. 
%
Qin et al.~\cite{qin2018bylabel} propose ByLabel as a new semi-automatic 
boundary-based image annotation tool, which is also presented as an 
improved alternative to LabelMe by Russell et al.~\cite{russell_labelme}. 
Bernal et al.~\cite{bernal2019gtcreator} presented GTCreator -- 
a stand-alone \emph{``flexible annotation tool for providing image and text 
annotations to image-based datasets.''}
Finally, for more complete and comprehensive surveys of the major tools and 
techniques for image annotation, we refer interested readers to 
Hanbury~\cite{hanbury2008survey} and 
Dasiopoulou et al.~\cite{dasiopoulou2011survey}.

Crowdsourcing is a fast and cost-effective way to make image annotation 
with a potentially large group of collaborators, i.e., annotators. 
Su et al.~\cite{su2012crowdsourcing} presented a system that collects 
bounding-box (i.e., bbox) annotations through crowdsourcing. 
Using crowdsourcing, however, creates a significant challenge on 
how to ensure that annotation quality and coverage is 
reasonably good~\cite{hanbury2008survey},~\cite{welinder2010online}. 
At the same time, crowdsourcing efforts need to pay particular attention to 
the reliability and differences between annotators during 
crowdsourcing~\cite{nowak2010reliable,su2012crowdsourcing,welinder2010online}.

\section{Detailed overview of the toolset}
\label{sec:detail}

The browser extension presented in this work offers a new way to annotate 
images, and the researchers can create their own dataset that includes images 
suitable for the exact need.


\subsection{Features}
\label{sec:toolfeatures}

We have been continuously designing, iterating, and improving the toolset's 
functionality. 
At present, \texttt{BRIMA} offers the following attractive and 
unique combination of features:

\begin{itemize}[leftmargin=*]
\itemsep0em

\item \textbf{Browser-only.} 
It does not require any installation or setup on the client-side, 
except of several clicks (e.g., ``Install Add-on'') when using most 
common and modern browsers.
Our tool allows to capture and annotate whatever
is currently displayed in the user browser. 
Unlike all current tools, this means there is no limitation what images 
are presented to the annotator. 
Therefore, it provides more diversity to the final dataset since the annotator 
can browse whatever URL or image s/he thinks is representative to the annotation 
task at hand, or according to the crowdsourcing instructions. 
In a nutshell, with our tool annotation is as simple as \emph{annotate-while-freely-browsing}.

\item \textbf{Web-oriented.} 
Traditional image annotation tools require that the images are first 
scraped from the web. 
Then, most often, they also require images to be organized by folders and filenames. 
Such an approach increases the complexity, the risk of errors, and the 
time required to pre-organize the dataset for annotation.
In our design, the annotation takes place within the extension and the browser.
This means that there is no need to scrape and organize the images 
to local files and folder, as the web images are already rendered within 
the extension which at the same time is the actual annotation tool. 
This also means there is no need to develop and maintain additional code 
that scrapes the web. 

\item \textbf{URL-aware.} 
Our tool by default can store, parse, and otherwise process URLs for various 
additional data, thanks to being run within the browser. 
For example, for Google Street View~\cite{anguelov2010google} it can 
parse almost exact location of the annotated object including additional 
parameters such as zoom and orientation of the Google Street View lens 
that captured the given view. 
However, it can be configured and used on virtually any set of websites, 
according to the needs of the crowdsourcing effort. 
Also, parsing routines can be added and customized separately on the basis of 
the domain/subdomain names, as well as prefixes and regular expressions matching 
certain URL patterns. 
This also means that there is always a possibility to trace-back where 
each training/validation/testing image or annotation was taken from.
Hence, this allows to further improve the annotation and data quality-check 
processes. 

\item \textbf{Minimal client-side.} 
Our browser extension does not require complex setup and installation on the \emph{client-side}. 
Everything can be done in browsers setup where the user just needs to 
configure/activate the extension. This makes large-scale crowdsourcing 
a breeze. 

\item \textbf{Minimal server-side.} 
It also requires minimal \emph{server-side} setup and configuration. 
We tested and release minimal setups supporting PHP and Python API backends. 

\item \textbf{Supports extensive combinations of ``OS + browser''.} 
The browser extension can be used in most common browsers and it has
been tested with multiple different combinations of OS and browsers 
(see Section~\ref{sec:tooleval}). 

\item \textbf{Support for standard/compatible data formats.} 
It supports natively MS COCO~\cite{lin2014microsoft} annotation JSON file formats, 
thus enabling immediate training of object detectors 
using a variety of CV/ML framework and backend combinations. 

\item \textbf{Out-of-the-box usage.} 
Our toolset can be used immediately and out-of-the-box by taking advantage 
of the provisioned client-side templates 
README documentation, and one-line scripts for server-side 
supporting HTTP/HTTPS with minimal software dependencies.

\item \textbf{Easy to configure.} 
Both the client-side browser extension and the server-side are easy to 
configure for various real-world scenarios. Most common browsers have 
a few click solution for activating browser extension and 
the server-side can be immediately started using a configurable 
command-line script. 

\item \textbf{Easy to use.} 
User interface is simple, and annotation is done with the mouse, 
with optional support for configurable keyboard shortcut keys. 
Color-coding helps to identify which object type/category the user is 
currently annotating (color-coding was noted by Gong et al.~\cite{gong1994image} as an important feature for annotation). 

\item \textbf{Configurable keyboard shortcuts.} 
Multiple keyboard shortcuts are defined, and can be customized. 
For example, a single key-press to select annotated object type or 
another key-press to accept all annotated object(s) 
and to send the full JSON data to the server backend.

\item \textbf{No pre-built dataset required.} 
This is a consequence of the \emph{web-oriented} design of our toolset, i.e., 
there is no need for pre-annotation web-scraping phase. 

\vspace{-4mm}
\end{itemize}

\subsection{Technical details}

Our entire toolset is composed of client-side (i.e., the browser extension), 
and the server-side which receives and stores the annotated data. 
The browser extension is written in pure JavaScript, and it is
helping the researchers and crowdsource contributors for fast and easy data annotation. 
For server-side, we provide basic reference implementations in PHP and 
Python (using Flask and Swagger/OpenAPI). 
However, nothing limits our toolset's users to rewrite the server-side in a 
different language and serve it from a different web-server setup.
The browser extension and the server communicate with each other via 
HTTP/XHR and RESTful API requests.

The exact annotation polygon data is based on the input from the user 
(typically using precise mouse clicks), and other parts of annotation data 
are automatically generated or enhanced by the browser extension. 
All the annotation data is sent to the server-side using a JSON format structure 
which at its core is compatible with the MS COCO and can be immediately used 
for CV/ML training under various object detection frameworks 
(e.g., TensorFlow~\cite{abadi_tensor}) and 
backends (e.g., CenterMask2~\cite{cm_git}, Detectron2~\cite{d2}).

\subsection{Toolset's workflow}

Our browser extension is easy to use. After an initial client-/server-side 
configuration, its usage follows a simple pattern:

\begin{enumerate}[leftmargin=*]
\itemsep0em

\item The user seeks an image from web resources such as photo stock 
websites (e.g., Flickr) or street view imagery (e.g., Google StreetView) 
corresponding to dataset collection needs. The key and novel idea is that 
annotator's browser viewport directly becomes the ``image to be annotated''. 

\item The user uses few keyboard/mouse actions to activate the image into our 
browser extension, and after that the selected viewport image is ready for annotation.

\item The user draws the annotation polygon around the object(s) of interest, 
and confirm object annotation.

\item The user, once all objects of interest are annotated, 
confirms the annotation is ready, and with a single 
click sends the data to the data collection server.

\item The server receives the data in MS COCO compatible format, processes it, 
and is ready to create a complete dataset snapshot ready for CV/ML training. 

\end{enumerate}

\begin{figure}[htb]
  \centering
  \includegraphics[width=0.75\columnwidth]{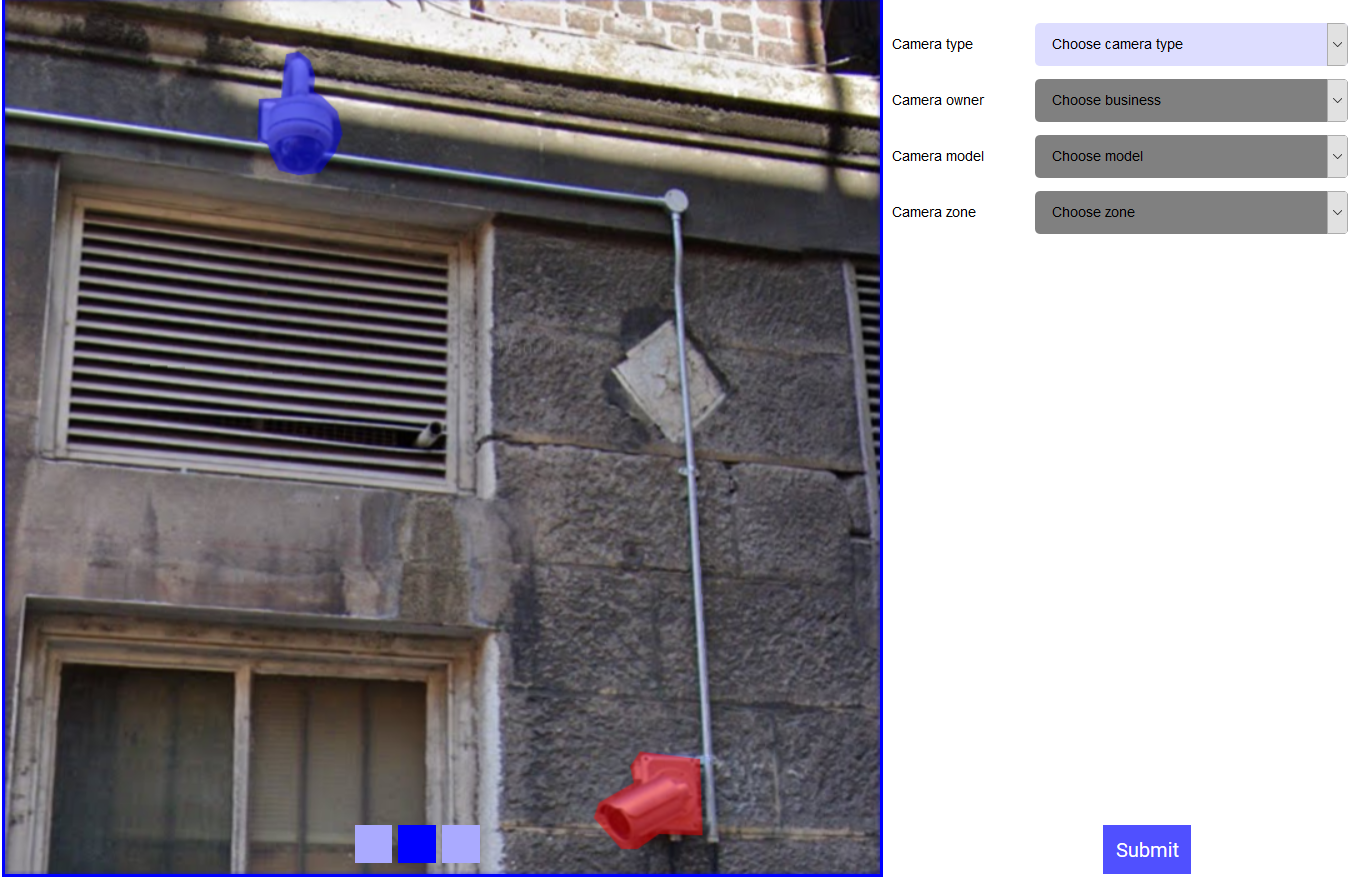}
  \caption{Browser window when extension is in annotation mode, 
  and the user annotated two CCTV cameras.
  }
  \label{fig:cctv}
\vspace{-4mm}
\end{figure}

\begin{figure}[htb]
  \centering
  \includegraphics[width=0.65\columnwidth]{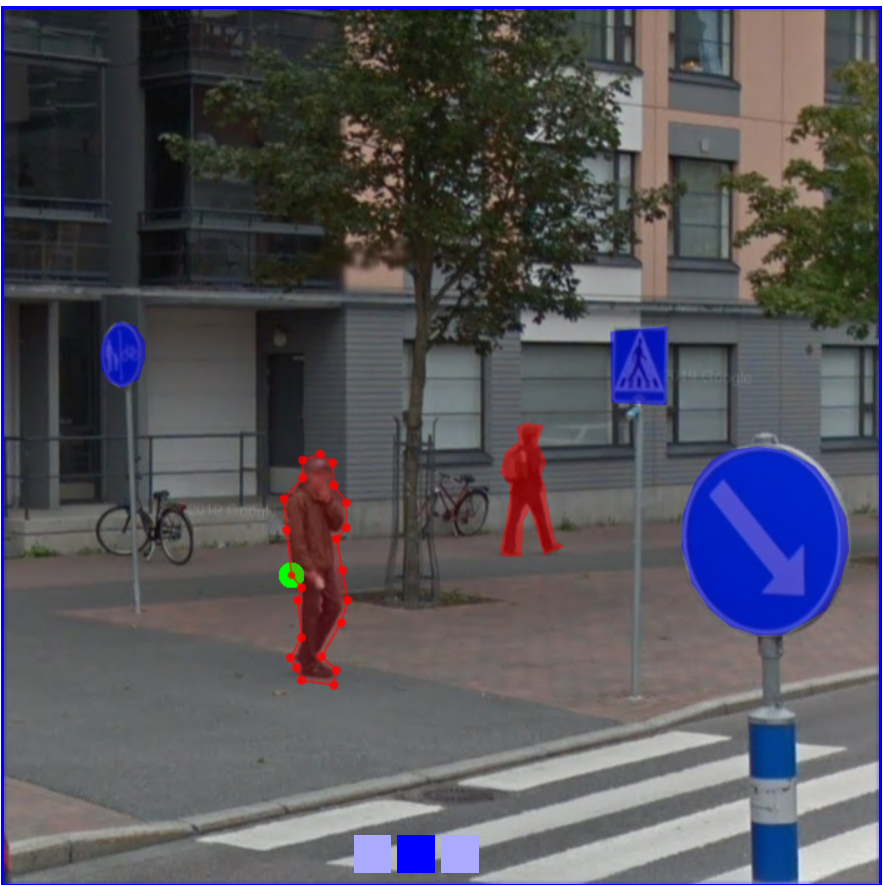}
  \caption{Browser extension can be used to annotate virtually any object of interest such as faces, humans, traffic signs.
  }
  \label{fig:other}
\vspace{-4mm}
\end{figure}

Figures~\ref{fig:cctv},~\ref{fig:other} present our tool in real-life action. 
In particular, Figure~\ref{fig:cctv} presents our extension used for 
\emph{privacy, anti-surveillance and safety applications}. 
In that particular setup we configured it to aid the collection of an annotated 
CCTV cameras dataset. We use such a dataset to build a state-of-the-art 
object detector for CCTV camera objects in real-life and in street-level imagery~\cite{turtiainen2020towards}. 
For the case of CCTV camera dataset collection, we easily configured the 
extension to allow the collection of additional data that is both specific and 
relevant for CCTV camera object. For example, Figure~\ref{fig:cctv} shows that 
the annotator can optionally fill the camera's exact model and 
camera's owner/operator. 
%
%
Alternatively, Figure~\ref{fig:other} presents our extension used for 
\emph{other privacy and generic Computer Vision applications}. 
For example, in ``other privacy'' applications it can be used to collect 
annotated datasets related to blurred/unblurred persons, faces, and license plate numbers.
Also, it can be used for general-purpose CV applications such as collecting 
accurate annotation of traffic signs and street infrastructure that can be 
subsequently used by self-driving cars and their autopilots.
%

\subsection{Toolset's validation}
\label{sec:tooleval}

We validate the effectiveness, efficiency, and usability of \texttt{BRIMA} 
toolset by actively employing it during a crowdsource effort 
aimed to create a large dataset used to train the 
\emph{first-ever state-of-the-art object detector for CCTV objects}~\cite{turtiainen2020towards}. 
Its purpose is to detect CCTV cameras in 
various images (e.g., street view) for privacy, anonymity, 
anti-surveillance, safety applications~\cite{lahtinen2021towards}. 
Such applications are important due to risks associated with 
lax cybersecurity of CCTV systems~\cite{costin2016security}, 
and their ``privacy invasion'' misuse once hacked. 

We recruited eight (8) image annotators which are mainly 
non-experts in crowdsourcing and image annotation. 
The entire crowdsourcing effort took place in September 2020 and overall, 
it took an equivalent of at least \emph{one and a half person-month effort}. 
Due to the ease of use and setup, and its \emph{web-oriented} and \emph{browser-only} 
design (see Section~\ref{sec:toolfeatures}), during the 72 hours experiment window 
the annotators were able to submit $\DSTwoTrainCountAll$ quality-checked images 
for a total of $\DSTwoTrainCountInstancesAll$ annotated instances of CCTV cameras 
($\DSTwoTrainCountDirected$ \emph{directed} and $\DSTwoTrainCountRound$ \emph{round}). 
%
Dataset results obtained using \texttt{BRIMA} 
are comparable with the state-of-the-art 
datasets of annotated images such as MS COCO where $6097$ is the median size of a 
training dataset for the particular object types such as cars, airplanes, stop signs. 
Thanks to the extension, the work and the crowdsourcing described in this paper, 
we were able to achieve state-of-the-art results in our wider efforts related 
to the CCTV camera object detector~\cite{turtiainen2020towards}. 
%

Table~\ref{tab:dataset2-images} presents overall summary of the clean and 
quality-checked dataset at the end of the crowdsource experiment. 
In Table~\ref{tab:dataset2-contrib} we present some detailed statistics 
for each crowdsourcing annotator 
and shortly explain each metric below:

\begin{itemize}[leftmargin=*]
\itemsep0em

\item \textbf{Approved Images.}
Number of images that successfully passed Quality Check (QC), both in terms of 
image and annotation. 

\item \textbf{DQ.}
DisQualification rate representing the percentage of images that were 
disqualified as a result of a failed QC. 
Hence the original number of submitted images per annotator is usually 
higher than the approved images. 
%

\item \textbf{Annotation expertise.}
Self-reporting scale was set from 0 (novice) to 5 (expert). 

\item \textbf{Easy setup and use.}
Self-reporting scale was set from 0 (extremely hard) to 5 (extremely easy). 

\item \textbf{Overall experience.}
Self-reporting scale was set from 0 (extremely bad) to 5 (extremely good). 

\item \textbf{OS and browser.}
The browser extension performed without any issues in a sufficiently diverse 
set of user environments (5 major OS releases and 3 major browser variants).

\vspace{-8mm}
\end{itemize}

\begin{table}[htb]
\centering
\caption{Crowdsourced dataset statistics.}
\label{tab:dataset2-images}
\resizebox{\columnwidth}{!}{%
    \begin{tabular}{l|r||l|r}%
    \toprule

    \multicolumn{2}{c}{Total counts} & \multicolumn{2}{c}{Instances grouped by sub-type} \\
    \midrule
    \specialcell{Total collected images}                           &   4167    & \specialcell{Directed camera instances}                        &   3325    \\
    \specialcell{Total annotated camera instances}                 &   5380    & \specialcell{Round camera instances}              &   2055    \\

    \midrule

    \multicolumn{2}{c}{Images grouped by source} & \multicolumn{2}{c}{Instances grouped by area} \\
    \midrule
    \specialcell{Google (Street View, Image Search)}                        &   3873    & \specialcell{Small (\textless 32x32 px)}                        &   1455    \\
    \specialcell{Baidu street view}              &   269    & \specialcell{Medium (32x32 -- 96x96 px)}              &   3345    \\
    \specialcell{Flickr}                 &   25    & \specialcell{Large (\textgreater 96x96 px)}                 &   580     \\

	\bottomrule
    \end{tabular}
}
\vspace{-6mm}
\end{table}

\begin{table}[htb]
\centering
\caption{Crowdsourcing statistics for each annotator.}
\label{tab:dataset2-contrib}
\resizebox{\columnwidth}{!}{%
    \begin{tabular}{l|r|r|c|r|r|r}%
    \toprule
    \specialcell{Annotator} & \specialcell{Approved \\ Images} & \specialcell{DQ \%} & \specialcell{OS and browser \\ (default 64-bit)} & \specialcell{Annotation \\ expertise \\ (0 -- 5)} & \specialcell{Easy setup \\ and use \\ (0 -- 5)} & \specialcell{Overall \\ experience \\ (0 -- 5)} \\
    \midrule
	\midrule
    Person1 JU & 418     &       1.7 \%    & \specialcell{MacOS (10.15.6) \\ FFox (80.0.1)}                 & 0      & 3.5      & 2      \\
    \midrule
    Person2 PK & 525     &       0.2 \%    & \specialcell{Debian (9.13) \\ Chromium (73.0.3683.75-dev)}     & 0     & 5     & 4      \\
    \midrule
    Person3 SK & 542     &       14.4 \%   & \specialcell{Win10 \\ FFox (81.0.1)}     & 3      & 5       & 4     \\
    \midrule
    Person4 AA & 228     &       15.4 \%   & no response     & no response      & no response        & no response      \\
    \midrule
    Person5 AT & 632     &       3.0 \%    & \specialcell{Win10 (1809) \\ FFox (78.3.0esr)}     & 1      & 3      & 4      \\
    \midrule
    Person6 TL & 750     &       0.7 \%    & \specialcell{Win10 \\ FFox (80.0.1)}     & 2      & 5        & 4      \\
    \midrule
    Person7 LS & 977     &       5.4 \%    & \specialcell{Ubuntu (18.04.5 LTS) \\ Chrome (85.0.4183.121)}     & 1      & 5     & 5      \\
    \midrule
    Person8 AC &  95     &       0.0 \%    & \specialcell{Ubuntu (16.04 LTS) \\ FFox (68.7.0esr) 32-bit}     & 4      & 5     & 5     \\
	\midrule
	\midrule
    \specialcell{ \underline{\textit{avg.}} \\ \textbf{sum} }      & \specialcell{ \underline{\textit{520.87}} \\ \textbf{4167} }      & \specialcell{ \underline{\textit{4.75 \%}} \\ -- }  & --        & \specialcell{ \underline{\textit{1.57}} \\ -- }     & \specialcell{ \underline{\textit{4.5}} \\ -- }    & \specialcell{ \underline{\textit{4}} \\ -- }    \\
	\bottomrule
    \end{tabular}
}
\vspace{-6mm}
\end{table}


\section{Conclusion}
\label{sec:concl}

We presented \texttt{BRIMA} -- the first browser-only annotation toolset that 
allows easy and fast image annotation within browser-viewport while incurring a low overhead overall. 
%
We also validated the effectiveness, efficiency, and usability of our 
extension/toolset by actively employing it during a successful crowdsource effort. 
This underpins the fast-to-bootstrap and easy-to-maintain characteristics 
of our toolset and the extension, therefore allowing the researchers to 
actually focus on the core of their experiments (e.g., training and 
experimenting state-of-the-art object detectors for various applications). 
We believe our toolset is a valuable contribution to the communities of 
researchers and practitioners, therefore the relevant artefacts 
(code, documentation, samples) will be made available open-source 
at: \url{https://github.com/tutalaht/brima}

\newpage

\section*{Acknowledgment}
Authors would like to acknowledge grants of computer capacity from the 
Finnish Grid and Cloud Infrastructure (FGCI) (persistent identifier 
\emph{urn:nbn:fi:research-infras-2016072533}).

Part of this research was kindly supported by the 
\emph{``Decision of the Research Dean on research funding within the Faculty (17.06.2020)''} 
grant from the Faculty of Information Technology of University of Jyv\"{a}skyl\"{a} 
(the grant was facilitated and managed by Dr. Andrei Costin). 

Hannu Turtiainen would like to also thank:
\begin{itemize}

\item The Finnish Cultural Foundation / Suomen Kulttuurirahasto (https://skr.fi/en) 
for supporting his Ph.D. dissertation work and research (grant decision 00211119). 

\item The Faculty of Information Technology of University of Jyvaskyla (JYU), 
in particular Prof. Timo H\"{a}m\"{a}l\"{a}inen, for partly supporting his Ph.D. supervission at JYU in 2021.

\end{itemize}

\bibliographystyle{IEEEbib}
{
    \footnotesize
    \bibliography{thesis-hannu,thesis-tuomo}
}


\end{document}